\documentclass{article}
\usepackage{spconf,amsmath,graphicx,hyperref}
\usepackage{amsfonts} 

\title{SOFT GRAPH TRANSFORMER FOR MIMO DETECTION}
\name{Jiadong Hong\textsuperscript{\rm 1}, Lei Liu\textsuperscript{\rm 1}, Xinyu Bian\textsuperscript{\rm 2}, Wenjie Wang\textsuperscript{\rm 2}, Zhaoyang Zhang\textsuperscript{\rm 1} \thanks{This work was supported in part by the National Natural Science Foundation of China (NSFC) under Grants 62301485, in part by the Zhejiang Provincial Natural Science Foundation under Grant LZ25F010002, and in part by the State Key Laboratory of Integrated Services Networks under Grant ISN25-10.}
}
\address{\textsuperscript{\rm 1}College of Information Science and Electronic Engineering, Zhejiang University, China, \\ \textsuperscript{\rm 2}Theory Lab, Central Research Institute, 2012 Labs, Huawei Technologies Co., Ltd., Hong Kong
}
\begin{document}
\ninept
\maketitle

\begin{abstract}
We propose the \emph{Soft Graph Transformer} (SGT), a soft-input–soft-output neural architecture designed for MIMO detection. While Maximum Likelihood (ML) detection achieves optimal accuracy, its exponential complexity makes it infeasible in large systems, and conventional message-passing algorithms rely on asymptotic assumptions that often fail in finite dimensions. Recent Transformer-based detectors show strong performance but typically overlook the MIMO factor graph structure and cannot exploit prior soft information. SGT addresses these limitations by combining \emph{self-attention}, which encodes contextual dependencies within symbol and constraint subgraphs, with \emph{graph-aware cross-attention}, which performs structured message passing across subgraphs. Its soft-input interface allows the integration of auxiliary priors, producing effective soft outputs while maintaining computational efficiency. Experiments demonstrate that SGT achieves near-ML performance and offers a flexible and interpretable framework for receiver systems that leverage soft priors.
\end{abstract}
\section{Introduction}

Multiple-Input Multiple-Output (MIMO) systems are fundamental to modern wireless communications, offering high spectral efficiency and robust links. Yet, efficient symbol detection remains challenging. Maximum Likelihood (ML) detection achieves optimal accuracy but is computationally prohibitive for large systems. Low-complexity message-passing methods such as Approximate Message Passing (AMP)~\cite{Donoho2009amp}, Orthogonal AMP (OAMP)~\cite{ma2017oamp}, and Memory AMP (MAMP)~\cite{liu2021mamp} alleviate complexity but are fragile due to their reliance on asymptotic assumptions. To improve robustness in finite-dimensional settings, deep-unfolded methods such as OAMPNet~\cite{he2018oampnet, he2020oampnet2} and DetNet~\cite{samuel2017deep} learn algorithmic parameters directly from data, thereby enhancing adaptability in finite-dimensional channels.

Recently, Transformer-based architectures~\cite{vaswani2017attention} have attracted attention in MIMO detection for their strong contextual understanding capability. RE-MIMO~\cite{pratik2020remimo} integrates recurrent networks with self-attention and achieves strong performance in correlated channels, but without explicit graph awareness. Transformer-based MIMO detector~\cite{ahmed2025transformer} employs QR decomposition for dimensional alignment, effective yet costly and overlooking the factor graph structure. By contrast, in channel decoding, Transformer-based designs explicitly leverage graph structures, e.g., ECCT~\cite{yoni2022ecct} aligns attention with the LDPC Tanner graph, and CrossMPT~\cite{park2025crossmpt} mimics message passing via cross-attention. These works suggest that structured attention can provide both efficiency and interpretability.

Inspired by classical message passing MIMO detection, we propose the \emph{Soft Graph Transformer} (SGT), a soft-input soft-output MIMO detector that inherits the self-attention module from Transformer-based MIMO~\cite{ahmed2025transformer} and the cross-attention module from CrossMPT~\cite{park2025crossmpt}, yet fundamentally differs from both. Unlike prior works where self- and cross-attention are employed in isolation, SGT integrates them within the AMP-style framework as factor-graph-guided mechanisms, thereby unifying contextual encoding and message passing in a principled manner. Specifically,

\begin{itemize}
\item Factor-graph-guided self- and cross-attention jointly realize contextual encoding and iterative message passing, distinguishing SGT from existing designs.
\item A soft-input soft-output interface naturally incorporates external priors from the other receiver blocks.
\end{itemize}

Experimental results show that SGT achieves near-ML detection accuracy as a standalone detector and consistently outperforms existing Transformer-based MIMO detectors. Moreover, its support for soft priors makes it a flexible building block for receiver architectures with auxiliary information.

\section{System Model}

\begin{figure*}[t]
    \centering
    \includegraphics[width=0.95\linewidth]{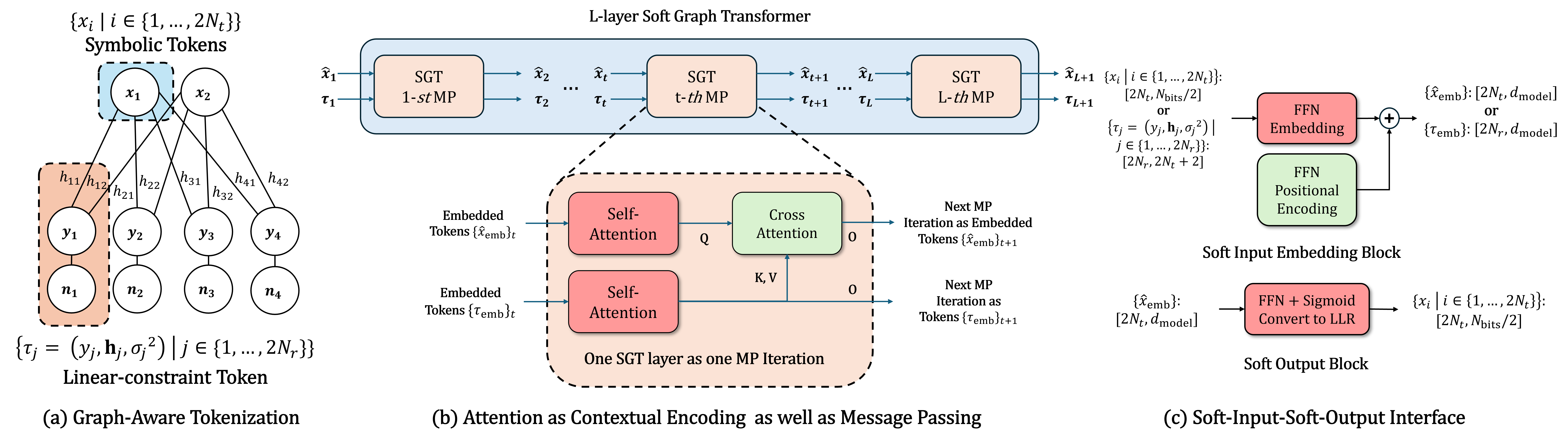}
    \caption{(a) Factor graph representation of the MIMO system, 
where $\mathbf{h}_j$ is the $j$-th row of the channel matrix $\mathbf{H}$, 
describing the connection between transmit symbols $\mathbf{x}$ and received signals $y_j$. 
(b) Unfolded SGT-based message passing (MP) detector with $L$ layers, 
where each SGT layer corresponds to one MP iteration. 
(c) Internal structure of one SGT layer, consisting of self-attention, 
cross-attention, and FFN modules for embedding, positional encoding, and LLR output. $N_\text{bits}$ in (c) refers to the number of bits per constellation symbol.}
    \label{fig:network-arch}
\end{figure*}

We consider a standard Multiple-Input Multiple-Output (MIMO) communication system, assuming perfect channel state information (CSI) at the receiver. The transmitter first encodes the information bits using an error correction code and maps the encoded bits to complex constellation symbols. These symbols are arranged into a time-frequency resource grid and transmitted across multiple transmit antennas.

At the receiver side, the channel model reduces to a stacked MIMO channel. In the complex domain, the received signal is expressed as
\begin{equation}
    \mathbf{y}_\mathbb{C} = \mathbf{H}_\mathbb{C} \mathbf{x}_\mathbb{C} + \mathbf{n}_\mathbb{C},
\end{equation}
where $\mathbf{y}_\mathbb{C} \in \mathbb{C}^{N_r}$ is the received signal vector, $\mathbf{x}_\mathbb{C} \in \mathbb{C}^{N_t}$ is the transmitted symbol vector, $\mathbf{H}_\mathbb{C} \in \mathbb{C}^{N_r \times N_t}$ is the channel matrix, and $\mathbf{n}_\mathbb{C} \sim \mathcal{CN}(0, \sigma_\text{c}^2 \mathbf{I})$ denotes additive white Gaussian noise.

For subsequent analysis, we adopt an equivalent \emph{real-valued representation} of the above complex system. Specifically, by separating the real and imaginary parts, the model can be rewritten as
\begin{equation}
    \underbrace{\begin{bmatrix}
        \Re(\mathbf{y}_\mathbb{C}) \\[2pt]
        \Im(\mathbf{y}_\mathbb{C})
    \end{bmatrix}}_{\mathbf{y} \in \mathbb{R}^{2N_r}}
    =
    \underbrace{\begin{bmatrix}
        \Re(\mathbf{H}_\mathbb{C}) & -\Im(\mathbf{H}_\mathbb{C}) \\
        \Im(\mathbf{H}_\mathbb{C}) & \Re(\mathbf{H}_\mathbb{C})
    \end{bmatrix}}_{\mathbf{H} \in \mathbb{R}^{2N_r \times 2N_t}}
    \underbrace{\begin{bmatrix}
        \Re(\mathbf{x}_\mathbb{C}) \\[2pt]
        \Im(\mathbf{x}_\mathbb{C})
    \end{bmatrix}}_{\mathbf{x} \in \mathbb{R}^{2N_t}}
    +
    \underbrace{\begin{bmatrix}
        \Re(\mathbf{n}_\mathbb{C}) \\[2pt]
        \Im(\mathbf{n}_\mathbb{C})
    \end{bmatrix}}_{\mathbf{n} \in \mathbb{R}^{2N_r}},
    \label{eq:mimosys}
\end{equation}
where $\mathbf{n} \sim \mathcal{N}(0, \tfrac{\sigma_\text{c}^2}{2}\mathbf{I}) = \mathcal{N}(\mathbf{0}, \boldsymbol{\Sigma})$ is real-valued Gaussian noise, \(\boldsymbol{\Sigma} = \mathrm{diag}(\sigma_1^2, \ldots, \sigma_{2N_r}^2)\) All subsequent analysis and algorithm design will be based on this real-valued model.

The receiver first uses an equalizer that enforces the linear model $\mathbf{y}=\mathbf{H}\mathbf{x}$ to regress estimates of the transmitted symbols $\mathbf{x}$. A symbol demapper then performs Bayesian denoising by exploiting the distribution of $\mathbf{x}$ for symbol-to-bit conversion. 

In AMP–based schemes, the equalizer can be viewed as a node that enforces the linear relation $\mathbf{y}=\mathbf{H}\mathbf{x}$, while the demapper acts as a node that models the distribution of the transmitted symbols. By iteratively exchanging soft information (also called message passing) between these two nodes, the receiver gradually refines its estimates, leading to improved detection performance.

\section{Methodology}

\subsection{Model Overview}

The proposed \emph{Soft Graph Transformer} (SGT) is a modular neural architecture for soft-input soft-output (SISO) detection in structured linear noisy systems. As illustrated in Fig.~\ref{fig:network-arch}, SGT processes probabilistic bit-level inputs and raw received features, updates them through a graph-informed Transformer backbone, and outputs bit-wise posterior likelihoods suitable for channel decoders.

The full inference pipeline consists of three main stages:

\begin{itemize}
    \item \textbf{Graph-Aware Tokenization (Fig.~\ref{fig:network-arch}(a)):} Constructs a bipartite graph from soft symbol estimates (Symbolic Tokens/Subgraph) and real-valued received features (Linear Constraint Tokens/Subgraph). 
    This tokenization step transforms the input data into a format suitable for the Transformer network, representing the relationships between transmitted symbols and received observations.

    \item \textbf{Attention as Contextual Encoding as well as MP (Fig.~\ref{fig:network-arch}(b)):} At its core, SGT applies a multi-layer Transformer block. Within each layer, a combination of self-attention and cross-attention operations is used. The self-attention is designed for better contextual token encoding, while the cross-attention mechanism is specifically designed to facilitate message passing from the linear constraint tokens to the symbol prior tokens, effectively enabling information exchange for iterative refinement.

    \item \textbf{Soft-Input Soft-Output Interface (Fig.~\ref{fig:network-arch}(c)):} Handles the processing of soft inputs and the generation of soft outputs. The Soft Input Embedding block takes prior LLRs and raw received features, converts them into embeddings, and combines them with positional encodings. The final token representations are then passed to the Soft Output Block, which maps them to posterior LLRs, ensuring compatibility with soft-input decoders.
\end{itemize}


\subsection{Graph-Aware Tokenization}
\label{sec:tokenization}

Unlike sparse Tanner graphs used in classical decoding, the factor graph induced by a MIMO system is inherently dense and weighted, where each edge encodes both dependency and a real-valued channel coefficient. 

To model the MIMO system in Eq.~\ref{eq:mimosys} with a graph representation suitable for Transformer design, 
we partition the factor graph into two subgraphs and represent their nodes as tokens:

\begin{itemize}
    \item \textbf{Linear-Constraint Tokens/Subgraph:} 
    Each observation $y_j$ corresponds to a factor node associated with the linear constraint 
    $y_j = \mathbf{h}_j \mathbf{x} + n_j$. 
    We tokenize these factor nodes into \emph{linear-constraint tokens} 
    \begin{equation}
        \mathcal{T}_{\text{lin}} = 
        \left\{ \tau_j = (y_j, \mathbf{h}_j, \sigma_j^2) \;\big|\; j \in \{1,\dots,2N_r\} \right\},
        \label{eq:linear-token-set}
    \end{equation}
    where $\mathbf{h}_j$ is the $j$-th row of $\mathbf{H}$ and $\sigma_j^2$ the noise variance, 
    encoding the local likelihood constraint between received signals and transmit symbols.

    \item \textbf{Symbolic Tokens/Subgraph:} 
    Each variable node corresponds to a transmit symbol $x_i$, 
    We tokenize these variable nodes into \emph{symbol prior tokens}
    \begin{equation}
        \mathcal{T}_{\text{sym}} = 
        \left\{ x_i \;\big|\; i \in \{1,\dots,2N_t\} \right\},
        \label{eq:symbol-token-set}
    \end{equation}
    serving as query embeddings that interact with constraint tokens through cross-attention.
\end{itemize}

This dual-token representation naturally aligns with the two subgraphs of the MIMO factor graph, 
where self-attention captures intra-subgraph consistency and cross-attention enables 
directed message passing between them.

As illustrated in Fig.~\ref{fig:network-arch}(a), this tokenization absorbs the channel state information (CSI) into node features, effectively transforming the weighted factor graph into a uniform-edge bipartite graph. This avoids the need for geometric interpretation of edges, a common challenge in Graph Transformer designs. Compared to Transformer-based MIMO, which compresses \((y_i, \mathbf{h}_i)\) via QR decomposition, our method preserves more information and yields better performance, as shown in ablation studies.

\subsection{Attention as Contextual Encoding and Message Passing}
\label{sec:message-passing}

\begin{figure}[t]
    \centering
    \includegraphics[width=0.7\linewidth]{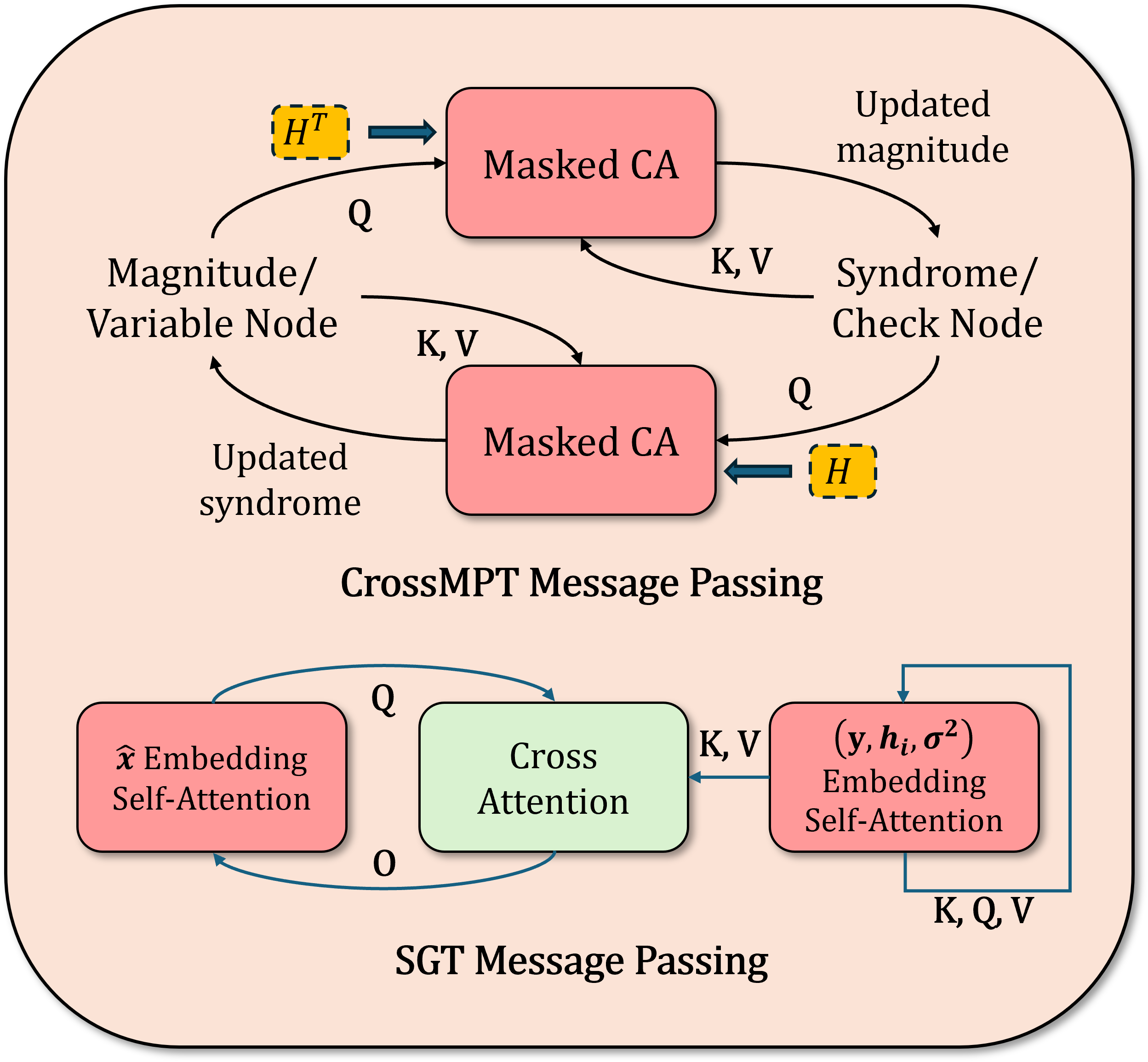}
    \caption{Comparison between CrossMPT and SGT message passing architectures. 
For decoding, constraints are defined between heterogeneous nodes 
(variable and check nodes) via the parity-check matrix $\mathbf{H}$, 
making cross-attention a natural choice as in CrossMPT. 
For MIMO, constraints are distributed across two isomorphic subgraphs: 
the linear constraint subgraph $\mathcal{T}_{\text{lin}}$ 
and the symbolic subgraph of symbol estimates $\mathcal{T}_{\text{sym}}$. 
Accordingly, SGT integrates \emph{self-attention} to encode contextual consistency within each subgraph 
and \emph{cross-attention} to exchange messages between them, 
thereby unifying contextual encoding with constraint-driven message passing.}
    \label{fig:sgt-mp}
\end{figure}

The attention mechanism for structured graph problems like MIMO detection serves two complementary purposes:

\textbf{Self-attention} provides powerful contextual encoding within homogeneous token sets, 
such as symbol estimates or linear-constraint features, ensuring consistency across similar entities. 
Formally, given a token set $\{\mathbf{t}_j\}_{j=1}^N$, the $j$-th updated token is obtained as:
\begin{equation}
    \tilde{\mathbf{t}}_j = \sum_{k=1}^N \alpha_{jk}\,\mathbf{W}_V \mathbf{t}_k, 
    \quad
    \alpha_{jk} = \text{softmax}\!\left(\frac{(\mathbf{W}_Q \mathbf{t}_j)^\top (\mathbf{W}_K \mathbf{t}_k)}{\sqrt{d_k}}\right),
\end{equation}
where $\mathbf{W}_Q, \mathbf{W}_K, \mathbf{W}_V$ are learnable projections. 
This mechanism encodes the global context of each subgraph.  

\textbf{Cross-attention} implements directed message passing across heterogeneous token types. 
For a query token $\mathbf{t}_j$ from one subgraph (e.g., symbol tokens) 
and a key-value token $\mathbf{t}_i$ from another subgraph (e.g., constraint tokens), the update is:
\begin{equation}
    \tilde{\mathbf{t}}_j = \sum_{i} \alpha_{ij}\,\mathbf{W}_V \mathbf{t}_i,
    \quad 
    \alpha_{ij} = \text{softmax}\!\left(\frac{(\mathbf{W}_Q \mathbf{t}_j)^\top (\mathbf{W}_K \mathbf{t}_i)}{\sqrt{d_k}}\right),
\end{equation}
where the message direction is determined by the roles of query and key tokens.  

The structural difference between \textbf{channel decoding} and \textbf{MIMO detection} 
guides how these mechanisms are applied. 
In channel decoding, \textbf{the coding constraint is naturally defined between two heterogeneous node types 
(variable nodes and check nodes).} 
Thus, CrossMPT~\cite{park2025crossmpt}, inspired by belief propagation (BP), 
primarily relies on cross-attention to mimic message updates across the Tanner graph. 

In contrast, MIMO detection constraints lie on two isomorphic subgraphs, respectively: 
(i) a \textbf{linear-constraint subgraph} defined by Eq.~\ref{eq:linear-token-set} and 
(ii) a \textbf{symbolic subgraph} defined by symbol beliefs defined by Eq.~\ref{eq:symbol-token-set}. 
Hence, the proposed SGT architecture naturally combines 
self-attention within each subgraph for contextual encoding and 
cross-attention between subgraphs for directional message passing.  
This dual use of attention mechanisms unifies AMP-inspired updates with Transformer architectures, 
as illustrated in Fig.~\ref{fig:sgt-mp}.

\begin{figure*}[t]
    \centering
    \includegraphics[width=0.95\linewidth]{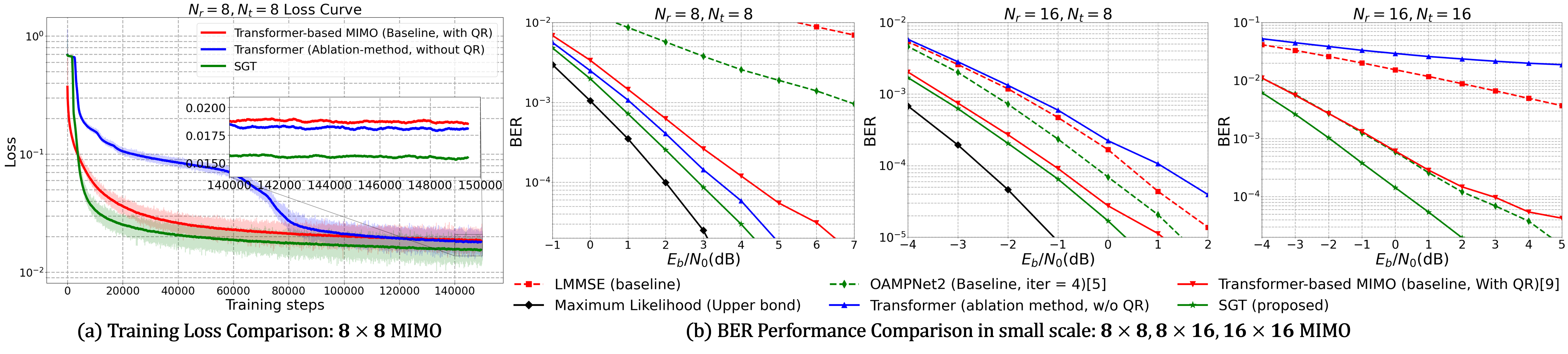}
    \caption{(a) Training Loss of 8$\times$8 MIMO (b) BER performance of 8$\times$8, 8$\times$16, 16$\times$16 Rayleigh Fading MIMO channel settings, SGT consistently outperforms Deep-Unfolded method OAMPNet2~\cite{he2020oampnet2} and Transformer-based MIMO~\cite{ahmed2025transformer}.}
    \label{fig:ber-standalone}
\end{figure*}

\subsection{Soft-Input–Soft-Output Interface}

In practical receivers, MIMO detectors often operate with soft priors provided by channel decoders or other auxiliary modules. To support iterative and modular receiver design, SGT is equipped with a soft-input–soft-output (SISO) interface that processes and outputs probabilistic information in a unified framework.

Specifically, SGT accepts two types of soft-valued inputs:
\begin{itemize}
    \item Prior symbol beliefs as symbolic tokens $\mathcal{T}_{\text{sym}}$;
    \item Channel observations as linear-constraint tokens $\mathcal{T}_{\text{lin}}$.
\end{itemize}

These inputs are independently embedded and mapped into a shared latent space, where positional encodings are applied before being processed by the stacked Transformer layers.

At the output, the refined symbolic embeddings are projected back to bit-level posterior likelihoods through a lightweight feed-forward network, producing soft outputs compatible with downstream decoders. This design enables end-to-end soft information processing, where priors are iteratively refined through graph-aware attention mechanisms.

By explicitly supporting soft inputs and outputs, SGT can be seamlessly integrated with SISO decoders or other detection modules, making it a flexible building block for iterative receiver architectures.


\section{Experiment}

\subsection{Ablation Studies}

We assess the effectiveness of graph-aware tokenization and cross-attention in standalone MIMO detection under Perfect-CSI Rayleigh fading channels with Quadrature Phase-Shift Keying (QPSK) modulation. Training and model settings is shown in Tab.~\ref{tab:simulation_config}. The compared methods include:

\begin{itemize}
\item \textbf{Baseline}: Encoder-only Transformer with QR preprocessing~\cite{ahmed2025transformer}.
\item \textbf{Ours w/o Cross-Attention}: Full graph-aware tokenization without QR; encoder-only; final FFN compresses embeddings from $2N_r$ to $2N_t$.
\item \textbf{Soft Graph Transformer (SGT)}: Full graph-aware tokenization with alternating self- and cross-attention layers.
\end{itemize}

\begin{table}[t]
\small
\centering
\caption{Settings for Ablation Studies}
\label{tab:simulation_config}
\begin{tabular}{l|c}
\hline
\textbf{Parameter} & \textbf{Value} \\
\hline
Number of SGT layers ($L$) & 8 \\
Number of attention heads & 8 \\
Model dimension ($d_{\text{model}}$) & 128 \\
Feed-forward  & 2-layer, ReLU, 128 hidden units. \\
Dropout rate & 0.1 \\
Constellation type & QPSK \\
Training SNR range & $[-5,\,15]$ dB \\
Number of training epochs & 250{,}000 \\
Optimizer & AdamW \\
Learning rate & $1\times10^{-4}$ \\
Training batch size & 500 ($16\times16$), 2500(else) \\
\hline
\end{tabular}
\end{table}

\noindent\textbf{Graph-Aware Tokenization: Better Information Preservation}

As shown in Fig.~\ref {fig:ber-standalone}(a), full tokenization without QR leads to lower final loss on small-scale systems (e.g., $8{\times}8$), validating its ability to preserve detailed symbol-level information. However, as system dimensions grow (e.g., $8{\times}16$, $16{\times}16$), training stagnates and performance degrades. Although Tokenization has advantages, it lacks the enhancement of data structure similar to QR preprocessing, and the training performance of the ablation scheme will be inferior to Transformer-based MIMO in relatively larger systems. This trend is reflected in the BER performance shown in Fig.~\ref {fig:ber-standalone}(b).

\noindent\textbf{Cross-Attention: Faster and Better Convergence}

SGT mitigates these limitations by introducing cross-attention, which facilitates explicit and scalable message passing between receive and transmit nodes. This mechanism serves a role analogous to QR pre-processing—guiding the learning process—while being fully learnable and end-to-end. As a result, SGT achieves both faster convergence (Fig.~\ref{fig:ber-standalone}(a)) and superior final accuracy (Fig.~\ref{fig:ber-standalone}(b)). The results highlight that both components—full tokenization and cross-attention—are indispensable: the former ensures information completeness, while the latter enables effective utilization.

\begin{table}[t]
\small
    \centering
    \caption{Runtime on GPU with 24GB memory and 16384 computing cores per 1000 samples (in seconds)}
    \label{tab:runtime}
    \begin{tabular}{lccc}
        \hline
        \textbf{Method} & \textbf{8$\times$8} & \textbf{8$\times$16} & \textbf{16$\times$16} \\
        \hline
        LMMSE                        & 0.00679  & 0.00718  & 0.00742  \\
        OAMP                         & 0.02208  & 0.02234  & 0.02408  \\
        OAMPNet2                     & 0.03333  & 0.03415  & 0.03507  \\
        Maximum Likelihood     & 2.10082  & 2.13612  & --       \\
        Transformer-based MIMO       & 0.03844  & 0.03924  & 0.04028  \\
        Transformer (Ablation)            & 0.03560  & 0.03494  & 0.03593  \\
        SGT (Proposed) & 0.09351  & 0.09464  & 0.09498  \\
        \hline
    \end{tabular}
\end{table}

\begin{table}[t]
\small
\centering
\caption{Comparison of MIMO Detection Algorithms in Terms of Computational Complexity. Transformer-based MIMO exhibits cubic complexity due to QR-decomposition pre-processing, while OAMP/OAMPNet also shows cubic complexity due to matrix inversion operations.}
\label{tab:complexity}
\begin{tabular}{l|c}
\hline
\textbf{Method} & \textbf{Complexity} \\
\hline
ML Detection & $O(M^{N_t}),\; M = 2^{N_\text{bits}}$ \\
OAMP / OAMPNet & $O(K N_r N_t^2)$ \\
Transformer-based MIMO & $O(N_r N_t^2 + L N_t^2 d_{\text{model}})$ \\
Proposed SGT (Ours) & $L \cdot O(N_r^2 + N_t^2 + N_r N_t)\cdot d_{\text{model}}$ \\
\hline
\end{tabular}
\end{table}

\subsection{Complexity Analysis}

\noindent\textbf{Small-Scale MIMO System:} For small settings, runtime results in Table~\ref{tab:runtime} show that SGT maintains runtime comparable to OAMPNet and Transformer-based MIMO, while delivering superior BER performance (Fig.~\ref{fig:ber-standalone}). This suggests that SGT achieves the best trade-off between complexity and performance.

\noindent\textbf{Large-scale MIMO System:} We analyze the online inference complexity of representative MIMO detection algorithms in Tab.~\ref{tab:complexity}. Let $d_\text{model}$ be the Transformer embedding dimension, $N_\text{bit}$ the number of bits per symbol, and $L$ the number of encoder layers. In ultra-large-scale systems, SGT exhibits a more favorable quadratic complexity growth. However, the detection performance under extremely large MIMO settings warrants further investigation.

\section{Conclusion}

In this work, we proposed the \emph{Soft Graph Transformer} (SGT), a soft-input–soft-output MIMO detector that unifies the contextual modeling ability of self-attention with the structured message passing of factor graphs. Unlike conventional detectors that either suffer from exponential complexity (ML) or fragility in finite dimensions (AMP, OAMP, MAMP), and unlike prior Transformer-based approaches that overlook the graph structure, SGT explicitly embeds factor-graph awareness into its attention mechanism. This design not only achieves near-ML detection accuracy in small-scale MIMO but also demonstrates superior robustness and efficiency compared to existing deep-unfolded and Transformer-based detectors. Furthermore, through its soft-input–soft-output interface, SGT seamlessly integrates external priors and has the flexibility to incorporate with various receiver blocks. These results highlight the promise of structured attention as a principled and interpretable foundation for next-generation deep-learning-based MIMO detection.

\label{sec:refs}

\bibliographystyle{IEEEbib}
\bibliography{refs}

@article{yoni2022ecct,
  title={Error Correction Code Transformer},
  author={Choukroun, Yoni and Wolf, Lior},
  journal={Advances in Neural Information Processing Systems},
  volume={35},
  pages={38695--38705},
  year={2022}
}

@inproceedings{
park2025crossmpt,
title={Cross{MPT}: Cross-attention Message-passing Transformer for Error Correcting Codes},
author={Seong-Joon Park and Hee-Youl Kwak and Sang-Hyo Kim and Yongjune Kim and Jong-Seon No},
booktitle={The Thirteenth International Conference on Learning Representations},
year={2025},
url={https://openreview.net/forum?id=gFvRRCnQvX}
}

@article{ahmed2025transformer,
  title={Transformer Learning-based Efficient MIMO Detection Method},
  author={Ahmed, Saleem and Kim, Sooyoung and others},
  journal={Physical Communication},
  volume={70},
  pages={102637},
  year={2025},
  publisher={Elsevier}
}

@ARTICLE{liu2021mamp,
  author={Liu, Lei and Huang, Shunqi and Kurkoski, Brian M.},
  journal={IEEE Transactions on Information Theory}, 
  title={Memory AMP}, 
  year={2022},
  volume={68},
  number={12},
  pages={8015-8039},
  doi={10.1109/TIT.2022.3186166}}

@article{Donoho2009amp,
   title={Message-Passing Algorithms for Compressed Sensing},
   volume={106},
   ISSN={1091-6490},
   url={http://dx.doi.org/10.1073/pnas.0909892106},
   DOI={10.1073/pnas.0909892106},
   number={45},
   journal={Proceedings of the National Academy of Sciences},
   publisher={Proceedings of the National Academy of Sciences},
   author={Donoho, David L. and Maleki, Arian and Montanari, Andrea},
   year={2009},
   month=nov, pages={18914–18919} }

@article{ma2017oamp,
  title={Orthogonal AMP},
  author={Ma, Junjie and Ping, Li},
  journal={IEEE Access},
  volume={5},
  pages={2020--2033},
  year={2017},
  publisher={IEEE}
}

@inproceedings{samuel2017deep,
  title={Deep MIMO Detection},
  author={Samuel, Neev and Diskin, Tzvi and Wiesel, Ami},
  booktitle={2017 IEEE 18th International Workshop on Signal Processing Advances in Wireless Communications (SPAWC)},
  pages={1--5},
  year={2017},
  organization={IEEE}
}

@article{vaswani2017attention,
  title={Attention is All You Need},
  author={Vaswani, Ashish and Shazeer, Noam and Parmar, Niki and Uszkoreit, Jakob and Jones, Llion and Gomez, Aidan N and Kaiser, {\L}ukasz and Polosukhin, Illia},
  journal={Advances in Neural Information Processing Systems},
  volume={30},
  year={2017}
}

@article{pratik2020remimo,
  title={RE-MIMO: Recurrent and Permutation Equivariant Neural MIMO Detection},
  author={Pratik, Kumar and Rao, Bhaskar D and Welling, Max},
  journal={IEEE Transactions on Signal Processing},
  volume={69},
  pages={459--473},
  year={2020},
  publisher={IEEE}
}

@ARTICLE{he2020oampnet2,
  author={He, Hengtao and Wen, Chao-Kai and Jin, Shi and Li, Geoffrey Ye},
  journal={IEEE Transactions on Signal Processing}, 
  title={Model-Driven Deep Learning for MIMO Detection}, 
  year={2020},
  volume={68},
  number={},
  pages={1702-1715},
  doi={10.1109/TSP.2020.2976585}}

@inproceedings{he2018oampnet,
  title={A model-driven deep learning network for MIMO detection},
  author={He, Hengtao and Wen, Chao-Kai and Jin, Shi and Li, Geoffrey Ye},
  booktitle={2018 IEEE Global Conference on Signal and Information Processing (GlobalSIP)},
  pages={584--588},
  year={2018},
  organization={IEEE}
}

\end{document}